# Avian Influenza (H5N1) Expert System using Dempster-Shafer Theory

Andino Maseleno, Md. Mahmud Hasan
Department of Computer Science, Faculty of Science, Universiti Brunei Darussalam
Jalan Tungku Link, Gadong BE 1410, Negara Brunei Darussalam
E-mail: andinomaseleno@yahoo.com, mahmud.hasan@ubd.edu.bn

*Abstract*— **Based on Cumulative Number of Confirmed Human Cases of Avian Influenza (H5N1) Reported to World Health Organization (WHO) in the 2011 from 15 countries, Indonesia has the largest number death because Avian Influenza which 146 deaths. In this research, the researcher built an Avian Influenza (H5N1) Expert System for identifying avian influenza disease and displaying the result of identification process. In this paper, we describe five symptoms as major symptoms which include depression, combs, wattle, bluish face region, swollen face region, narrowness of eyes, and balance disorders. We use chicken as research object. Research location is in the Lampung Province, South Sumatera. The researcher reason to choose Lampung Province in South Sumatera on the basis that has a high poultry population. Dempster-Shafer theory to quantify the degree of belief as inference engine in expert system, our approach uses Dempster-Shafer theory to combine beliefs under conditions of uncertainty and ignorance, and allows quantitative measurement of the belief and plausibility in our identification result. The result reveal that Avian Influenza (H5N1) Expert System has successfully identified the existence of avian influenza and displaying the result of identification process.**

I. INTRODUCTION

Historically, outbreaks of avian influenza first occurred in Italy in the 1878, when it was a lot of dead birds [1]. Then came another outbreak of Avian Influenza in Scotland in the 1959 [2]. The virus that causes Avian Influenza in Italy and Scotland are the current strain of H5N1 virus appeared again attacked poultry and humans in various countries in Asia, including Indonesia, which caused many deaths in humans [3].

Avian influenza virus H5N1, which has been limited to poultry, now has spread to migrating birds and has emerged in mammals and among the human population. It presents a distinct threat of a pandemic for which the World Health Organization and other organizations are making preparations. In the 2005, the World Health Assembly urged its Member States to develop national preparedness plans for pandemic influenza [4]. Developing countries face particular planning and other challenges with pandemic preparedness as there may be a higher death rate in developing countries compared with more developed countries [5]. In this research, we use chicken as research object because chicken population has grown very fast in Lampung Province in the 2009, native chicken population around 11,234,890, broiler population around 15,879,617, layer population around 3,327,847. Lampung Province has been divided into 10 regencies, 204 districts and 2279 villages with area of 3,528,835 hectare [6].

Expert System is a computer application of artificial intelligence [7],[9],[10] which contains a knowledge base and an inference engine [8]. Since such programs attempt to emulate the thinking patterns of an expert, it is natural that the first work was done in artificial intelligence circles. Among the first expert systems were the 1965 Dendral programs [11], which determined molecular structure from mass spectrometer data; R1 [12] used to configure computer systems; and MYCIN [13] for medical diagnosis. The basic idea behind expert system is simply that expertise, which is the vast body of task-specific knowledge, is transferred from a human to a computer. This knowledge is then stored in the computer and users call upon the computer for specific advice as needed. The computer can make inferences and arrive at a specific conclusion. Then like a human consultant, it gives advices and explains, if necessary, the logic behind the advice [14]. Some expert systems for diagnosis in avian influenza have been developed which were expert system for avian influenza disease using Visual Basic.NET [15], and web based [16]. Actually, according to researchers knowledge, Dempster-Shafer theory of evidence has never been used for built an expert system for identifying avian influenza.

The remainder is organized as follows. The uncertainty in expert system is briefly reviewed in Section 2. Section 3 details the proposed Dempster-Shafer Theory. Architecture of Avian Influenza (H5N1) Expert System in Section 4. The experimental results are presented in Section 5, and final remarks are concluded in Section 6.

II. UNCERTAINTY IN EXPERT SYSTEM

The construction of expert and other intelligent computer systems requires sophisticated mechanism for representing and reasoning with uncertain information. At least three forms of uncertainty can be identified as playing a significant role in these types of systems. The first of these possibilistic uncertainty appears in situations where the value of a variable can only be narrowed down to a set of values one of which is the actual value of the variable. The second kind of uncertainty is related to situations in which there exists uncertainty as to satisfaction of a predicate by an element. This is manifested by concepts which have imprecise or gray boundaries. A very powerful tool for handling this type of uncertainty which also handles the first type of uncertainty is the fuzzy set. The third type of uncertainty is related to situations in which the value of a variable assumes can be modeled by the performance of a random experiment [17]. Figure 1 shows the process for dealing with uncertainty in Avian Influenza (H5N1) Expert System.

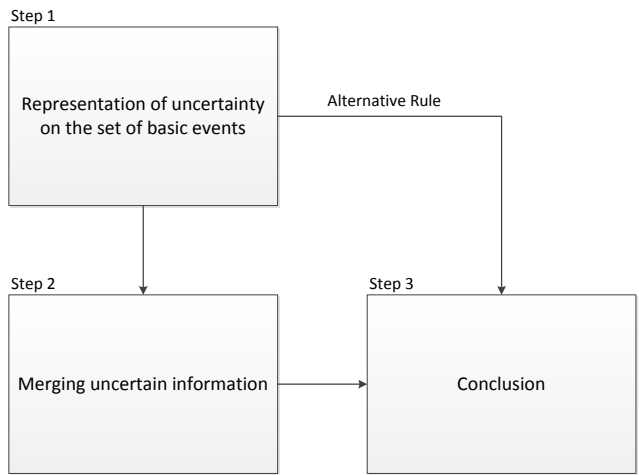

Figure 1. The process for dealing with uncertainty in Avian Influenza (H5N1) Expert System

In step 1, an expert provides uncertain knowledge in the form of rules with the possibility. These rules are probability values. In step 2, the uncertain knowledge on a set of basic events can be directly used to draw conclusions in simple cases (step 3). However, in many cases the various events associated with each other. Therefore, it is necessary to combine the information contained in step 1 into the global value system. In step 3, the goal is knowledge-based systems draw conclusions. It is derived from uncertain knowledge in steps 1 and 2, and is usually implemented by an inference engine. Working with the inference engine, the expert can adjust the input that they enter in Step 1 after displaying the results in steps 2 and 3.

## III. DEMPSTER-SHAFER THEORY

The Dempster-Shafer theory was first introduced by Dempster [18] and then extended by shafer [19], but the kind of reasoning the theory uses can be found as far back as the seventeenth century. This theory is actually an extension to classic probabilistic uncertainty modeling. Whereas the Bayesian theory requires probabilities for each question of interest, belief functions allow us to base degrees of belief for on question on probabilities for a related question. The advantages of the Dempster-Shafer theory as follows:
1. It has the ability to model information in a flexible way without requiring a probability to be assigned to each element in a set,
2. It provides a convenient and simple mechanism (Dempster's combination rule) for combining two or more pieces of evidence under certain conditions.
3. It can model ignorance explicitly.
4. Rejection of the law of additivity for belief in disjoint propositions.

Avian Influenza (H5N1) Expert System using Dempster-Shafer theory in the decision support process. Flowchart of Avian Influenza (H5N1) Expert System shown in Figure 1.

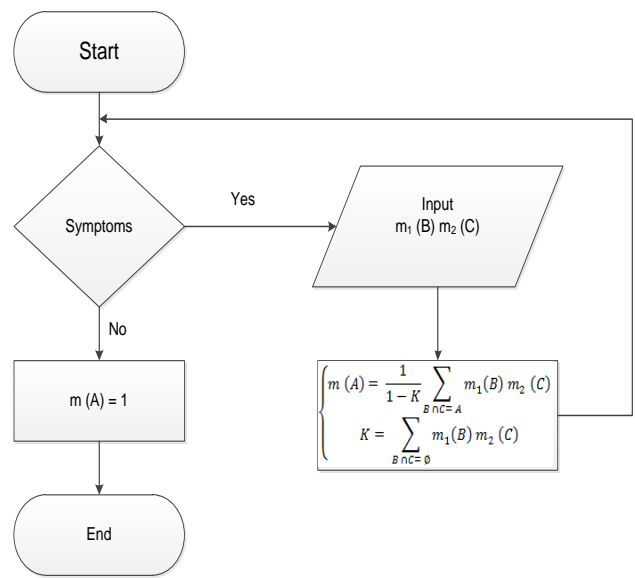

Figure 2. Flowchart of Avian Influenza (H5N1) Expert System

The consultation process begins with selecting the symptoms. If there are symptoms then will calculate, The Dempster-Shafer theory provides a rule to combine evidences from independent observers and into a single and more informative hint. Evidence theory is based on belief function and plausible reasoning. First of all, we must define a frame of discernment, indicated by the sign $\Theta$. The sign $2^\Theta$ indicates the set composed of all the subset generated by the frame of discernment. For a hypothesis set, denoted by A, m(A)→[0,1].

$m(\emptyset) = 0$

$$\sum_{A \in 2^\Theta} m(A) = 1 \qquad (1)$$

$\emptyset$ is the sign of an empty set. The function m is the basic probability assignment. Dempster's rule of combination combines two independent sets of mass assignments.

$(m_1 \oplus m_2)(\emptyset) = 0 \qquad (2)$

$$(m_1 \oplus m_2)(A) = \frac{1}{1-K} \sum_{B \cap C = A} m_1(B)\, m_2(C) \qquad (3)$$

Where

$$K = \sum_{B \cap C = \emptyset} m_1(B)\, m_2(C) \quad (4)$$

$$m(A), m_1(B), m_2(C) \to [0,1], A \neq \emptyset$$

The results are that Avian Influenza (H5N1) Expert System successfully identifying disease and displaying the result of identification process.

## IV. AVIAN INFLUENZA (H5N1) EXPERT SYSTEM

Avian Influenza (H5N1) Expert System has four main architectural components that are the knowledge base, the inference engine, the knowledge acquisition module, and the user interface for input/output. Figure 3 shows architecture of Avian Influenza (H5N1) Expert System.

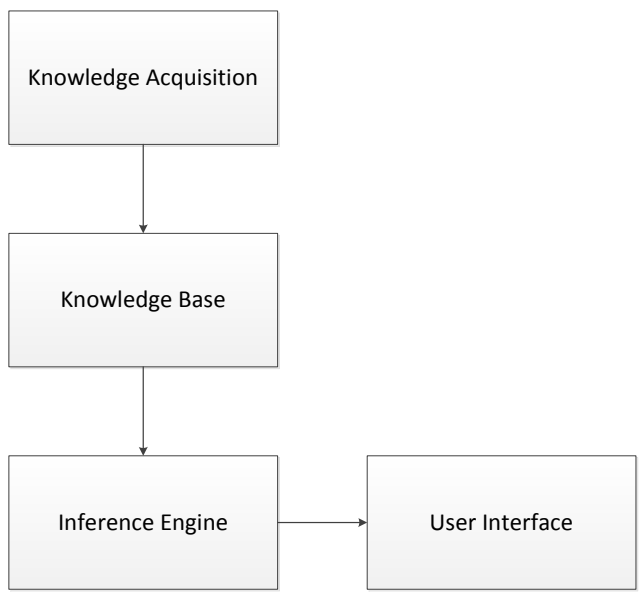

Figure 3. Architecture of Avian Influenza (H5N1) Expert System

### 4.1 Knowledge acquisition

In the present work, knowledge has been obtained from two sources. We acquired textual information from literature such as extension booklets, reports, papers, etc., related to the avian influenza diseases. The printed material allowed became familiar with the subject and a more effective communication with the experts. Most knowledge was acquired from the experts using conventional interviewing techniques. The interview methods permitted a better understanding of the problem and its further representation. This knowledge was provided by experts on avian influenza. Unstructured and structured interviews were used. The unstructured interviews were used to define the familiar tasks involved in the process of identification, to obtain an initial understanding of the range of complications involved, and to define specific problems for later discussion. The questions were more or less spontaneous and notes were taken on discussion. These methods were complemented with structured interviews. In the structured interviews, we revised and discussed in depth familiar tasks to clarify questions.

### 4.2 Knowledge Base

A critical aspect of building an expert system is formulating the scope of the problem and gleaning from the source expert the domain information needed to solve the problem. The reliability of an expert system depends on the quality of knowledge contained in the knowledge base. In this research, we built a Avian Influenza (H5N1) Expert System and we describe five symptoms which include depression, combs, wattle, bluish face region, swollen face region, narrowness of eyes, and balance disorders. Knowledge of which input on the testing of this expert system can be seen in Table I.

TABLE I. KNOWLEDGE BASE TABLE

| No. | Symptom | Disease | Basic Probability Assignment |
|---|---|---|---|
| 1 | Depression | Avian Influenza | 0.70 |
| | | Newcastle Disease | |
| | | Fowl Cholera | |
| | | Infectious Bronchitis respiratory form | |
| | | Infectious Bronchitis reproduction form | |
| | | Swollen Head Syndrome | |
| 2 | Combs, wattle, bluish face region | Avian Influenza | 0.90 |
| 3 | Swollen face region | Avian Influenza | 0.83 |
| | | Newcastle Disease | |
| | | Fowl Cholera | |
| 4 | Narrowness of eyes | Swollen Head Syndrome | 0.90 |
| 5 | Balance disorders | Newcastle Diseases | 0.60 |
| | | Swollen Head Syndrome | |

### 4.3 Inference Engine

The inference engine is charged with the role of performing and controlling inference on the knowledge base. Specific features of the inference engine depend on the knowledge representation scheme used for the knowledge base. Because the most common representation scheme is production rules, this inference engine will be exemplified. In this research we use forward chaining inference engine using Dempster-Shafer Theory. The following will be shown the inference engine using Dempster-Shafer Theory to diagnose avian influenza.

Symptoms:
1. Depression
2. Combs, wattle, bluish face region
3. Swollen face region
4. Narrowness of eyes
5. Balance disorder

A. *Symptom 1*

Depression is a symptom of Avian Influenza (AI), Newcastle Disease (ND), Fowl Cholera (FC), Infectious Bronchitis respiratory form (IBRespi), Infectious Bronchitis reproduction form (IBRepro), and Swollen Head Syndrome (SHS). The measures of uncertainty, taken collectively are known in Dempster-Shafer Theory terminology as a ``basic probability assignment'' (bpa).

Hence we have a bpa, say $m_1$ of 0.7 given to the focal element {AI, ND, FC, IBRespi, IBRepro, SHS} in example, $m_1$({AI, ND, FC, IBRespi, IBRepro, SHS}) = 0.7, since we know nothing about the remaining probability it is allocated to the whole of the frame of the discernment in example, $m_1$({AI, ND, FC, IBRespi, IBRepro, SHS}) = 0.3, so:
$m_1${AI, ND, FC, IBRespi, IBRepro, SHS} = 0.7
$m_1\{\Theta\}$ = 1 - 0.7 = 0.3 (5)

### B. Symptom 2

Combs, wattle, bluish face region are symptoms of Avian Influenza with a bpa of 0.9, so that:
$m_2${AI} = 0.9
$m_2\{\Theta\}$ = 1 − 0.9 = 0.1 (6)

With the symptoms comb, wattle, bluish face region then required to calculate the new bpa values for some combinations ($m_3$). Combination rules for the $m_3$ can be seen in the Table II.

TABLE II. COMBINATION OF SYMPTOM 1 AND SYMPTOM 2

|  |  | {AI} | 0.9 | Θ | 0.1 |
|---|---|---|---|---|---|
| {AI, ND, FC, IBRespi, IBRepro, SHS} | 0.7 | {AI} | 0.63 | {AI, ND, FC, IBRespi, IBRepro, SHS} | 0.07 |
| Θ | 0.3 | {AI} | 0.27 | Θ | 0.03 |

$m_3 (AI) = \dfrac{0.63 + 0.27}{1-0} = 0.9$

$m_3 (AI, ND, FC, IBRespi, IBRepro, SHS) = \dfrac{0.07}{1-0} = 0.07$

$m_3 (\Theta) = \dfrac{0.03}{1-0} = 0.03$ (7)

### C. Symptom 3

Swollen face region is a symptom of Avian Influenza, Newcastle Disease, Fowl Cholera with a bpa of 0.83, so that
$m_4$ {AI, ND, FC} = 0.83
$m_4 (\Theta)$ = 1 − 0.83 = 0.17 (8)

With the symptom of swollen face region then required to calculate the new bpa values for each subset. Combination rules for the $m_5$ can be seen in Table III.

TABLE III. COMBINATION OF SYMPTOM 1, SYMPTOM 2, AND SYMPTOM 3

|  |  | {AI, ND, FC} | 0.83 | Θ | 0.17 |
|---|---|---|---|---|---|
| {AI} | 0.9 | {AI} | 0.747 | {AI} | 0.153 |
| {AI, ND, FC, IBRespi, IBRepro, SHS} | 0.07 | {AI, ND, FC} | 0.0581 | {AI, ND, FC, IBRespi, IBRepro, SHS} | 0.0119 |
| Θ | 0.03 | {AI, ND, FC} | 0.0249 | Θ | 0.0051 |

$m_5 (AI) = \dfrac{0.747 + 0.153}{1-0} = 0.9$

$m_5 (AI, ND, FC) = \dfrac{0.0581 + 0.0249}{1-0} = 0.083$

$m_5 (AI, ND, FC, IBRespi, IBRepro, SHS) = \dfrac{0.0119}{1-0} = 0.0119$

$m_5 (\Theta) = \dfrac{0.0051}{1-0} = 0.0051$ (9)

### D. Symptom 4

Narrowness of eyes is a symptom of Swollen Head Syndrome with a bpa of 0.9, so that:
$m_6$ (SHS) = 0.9
$m_6 (\Theta)$ = 1 − 0.9 = 0.1 (10)

With narrowness of eyes in the absence of symptoms then required to recalculate the new bpa values for each subset with bpa $m_7$. Combination rules for $m_7$ can be seen in Table IV.

TABLE IV. COMBINATION OF SYMPTOM 1, SYMPTOM 2, SYMPTOM 3, AND SYMPTOM 4

|  |  | {SHS} | 0.9 | Θ | 0.1 |
|---|---|---|---|---|---|
| {AI} | 0.9 | Θ | 0.81 | {AI} | 0.09 |
| {AI, ND, FC} | 0.083 | Θ | 0.0747 | {AI, ND, FC} | 0.0083 |
| {AI, ND, FC, IBRespi, IBRepro, SHS } | 0.0119 | {SHS} | 0.01071 | {AI, ND, FC, IBRespi, IBRepro, SHS } | 0.00119 |
| Θ | 0.0051 | {SHS} | 0.00459 | Θ | 0.00051 |

$m_7 (SHS) = \dfrac{0.01071 + 0.00459}{1-(0.81+0.0747)} = 0.13270$

$m_7 (AI) = \dfrac{0.09}{1-(0.81+0.0747)} = 0.78057$

$m_7 (AI, ND, FC) = \dfrac{0.0083}{1-(0.81+0.0747)} = 0.07199$

$m_7$ (AI, ND, FC, IBRespi, IBRepro, SHS) = $\dfrac{0.00119}{1-(0.81+0.0747)} = 0.01032$

$m_7 (\Theta) = \dfrac{0.00051}{1-(0.81+0.0747)} = 0.00442$ (11)

### E. Symptom 5

Balance disorders is a symptom of Newcastle Diseases and Swollen Head Syndrome with a bpa of 0.6, so that:
$m_8$ {ND,SHS} = 0.6
$m_8 \{\Theta\}$ = 1 - 0.6 = 0.4 (12)

With balance disorders symptom will be required to recalculate the new bpa values for each subset with $m_9$ bpa. Combination rules for the $m_9$ can be seen in Table V.

TABLE V. COMBINATION OF SYMPTOM 1, SYMPTOM 2, SYMPTOM 3, SYMPTOM 4, AND SYMPTOM 5

|  |  | {ND, SHS} | 0.6 | Θ | 0.4 |
|---|---|---|---|---|---|
| {SHS} | 0.13270 | {SHS} | 0.07962 | {SHS} | 0.05308 |
| {AI} | 0.78057 | Θ | 0.46834 | {AI} | 0.31222 |
| {AI, ND, FC} | 0.07199 | {ND} | 0.04319 | {AI, ND, FC} | 0.02880 |
| {AI, ND, FC, IBRespi, IBRepro, SHS } | 0.01032 | {ND, SHS} | 0.00619 | {AI, ND, FC, IBRespi, IBRepro, SHS } | 0.00413 |
| Θ | 0.00442 | {ND, SHS} | 0.00265 | Θ | 0.00177 |

$$m_9 (SHS) = \frac{0.07962 + 0.05308}{1 - 0.46834} = 0.24960$$

$$m_9 (AI) = \frac{0.31222}{1 - 0.46834} = 0.58725$$

$$m_9 (ND) = \frac{0.04319}{1 - 0.46834} = 0.08124$$

$$m_9 (ND, SHS) = \frac{0.00619 + 0.00265}{1 - 0.46834} = 0.01663$$

$$m_9 (AI, ND, FC) = \frac{0.02880}{1 - 0.46834} = 0.05417$$

$$m_9 (AI, ND, FC, IBRespi, IBRepro, SHS) = \frac{0.00413}{1 - 0.46834} = 0.00777$$

$$m_9 (\Theta) = \frac{0.000232}{1 - 0.061038} = 0.00025 \quad (13)$$

The final result can be seen in Table VI.

TABLE VI. FINAL RESULT

| No. | Disease | Basic Probability Assignment |
|---|---|---|
| 1 | Swollen Head Syndrome | 0.24960 |
| 2 | Avian Influenza | 0.58725 |
| 3 | Newcastle Disease | 0.08124 |
| 4 | Newcastle Disease<br>Swollen Head Syndrome | 0.01663 |
| 5 | Avian Influenza<br>Newcastle Disease<br>Fowl Cholera | 0.05417 |
| 6 | Avian Influenza<br>Newcastle Disease<br>Fowl Cholera<br>Infectious Bronchitis respiratory form<br>Infectious Bronchitis reproduction form<br>Swollen Head Syndrome | 0.00777 |

The most highly bpa value is the $m_9$ (AI) that is equal to 0.58725 which means the possibility of a temporary diseases with symptoms of depression, comb, wattle, bluish face region, swollen region face, narrowness of eyes, and balance disorders is the Avian influenza (H5N1).

### 4.4 User Interface

A user interface is the method by which the expert system interacts with a user. The input or output interface defines the way in which the expert system interacts with the user and other systems such as databases. Interfaces are graphical with screen displays, windowing, and mouse control. They receive input from the user and display output to the user The user interface development of Avian Influenza (H5N1) Expert System to begin with designing web pages and designing a web-based applications. Designing web pages using PHP, the web page then connected to the MySQL database, after that designing a web-based applications which used to access the network database. The relationship between applications with network database has shaped overall applications to manage information to be displayed.

### V. IMPLEMENTATION

The following will be shown the working process of expert system in diagnosing a case. The consultation process begins with selecting the symptoms found on the list of symptoms. In the cases tested, a known symptoms are depression, comb, wattle, bluish-colored facade region, region of the face swollen, eyes narrowed and balance disorders. The consultation process and can be seen in Figure 4.

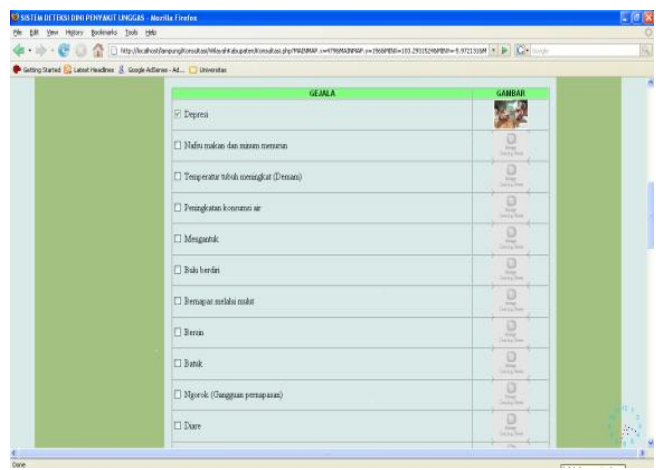

Figure 4. Symptoms Selection

In the case of depression, comb, wattle and region of the face bluish, region of the face swollen, eyes narrowed and lachrymal glands swollen. The result of consultation is avian influenza with bpa value equal to 0.587275693312. Figure 5 shows the result of consultation.

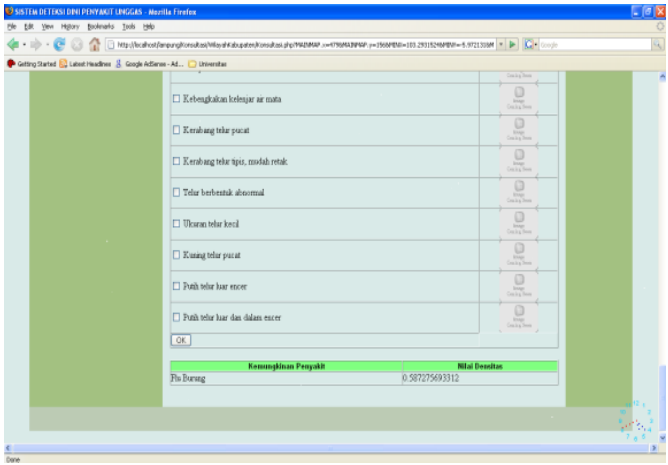

Figure 5. The Result of Consultation

## VI. CONCLUSION

Identification of avian influenza can be performed using Dempster-Shafer Theory. In this paper we describe five symptoms as major symptoms which include depression, combs, wattle, bluish face region, swollen face region, narrowness of eyes, and balance disorders. The simplest possible method for using probabilities to quantify the uncertainty in a database is that of attaching a probability to every member of a relation, and to use these values to provide the probability that a particular value is the correct answer to a particular query. An expert in providing knowledge is uncertain in the form of rules with the possibility, the rules are probability value. The knowledge is uncertain in the collection of basic events can be directly used to draw conclusions in simple cases, however, in many cases the various events associated with each other. Knowledge based is to draw conclusions, it is derived from uncertain knowledge. Reasoning under uncertainty that used some of mathematical expressions, gave them a different interpretation: each piece of evidence (finding) may support a subset containing several hypotheses. This is a generalization of the pure probabilistic framework in which every finding corresponds to a value of a variable (a single hypothesis). In this research, Avian Influenza (H5N1) Expert System has been successfully identifying Avian Influenza diseases and displaying the result of identification process This research can be an alternative in addition to direct consultation with doctor and to find out quickly avian influenza (H5N1) disease.